\documentclass[letterpaper, 10 pt, conference]{ieeeconf}  

\IEEEoverridecommandlockouts                              

\usepackage{amsmath}
\usepackage{amssymb} 
\usepackage{graphicx}
\usepackage{subcaption}
\usepackage{booktabs}
\usepackage{tabu}
\usepackage{tikz}
\usepackage{pgfplots}
\pgfplotsset{compat=1.18}
\usepackage{microtype}

\title{\LARGE \bf
Ground Plane Projection for Improved Traffic Analytics at Intersections
}

\author{Sajjad Pakdamansavoji$^{1}$, Kumar Vaibhav Jha$^{1}$, Baher Abdulhai$^{2}$, James H. Elder$^{1}$
\thanks{*This work was supported by the Center of Vision Research (CVR) and Centre for AI \& Society (CAIS) at York University, the Vector Institute for Artificial Intelligence, the Region of York and Trans-Plan.}
\thanks{$^{1}$Department of Electrical Engineering and Computer Science, York University, Toronto, ON, CA.
        {\tt\small sj.pakdaman.edu@gmail.com, kj323@yorku.ca, jelder@yorku.ca}}%
\thanks{$^{2}$Department of Civil and Mineral Engineering, University of Toronto, Toronto, ON, CA.
        {\tt\small baher.abdulhai@utoronto.ca}}%
}

\begin{document}

\maketitle
\thispagestyle{empty}
\pagestyle{empty}

\begin{abstract}
Accurate turning movement counts at intersections are important for signal control, traffic management and urban planning. Computer vision systems for automatic turning movement counts typically rely on visual analysis in the image plane of an infrastructure camera. Here we explore potential advantages of back-projecting  vehicles detected in one or more infrastructure cameras to the ground plane for analysis in real-world 3D coordinates.  For single-camera systems we find that back-projection yields more accurate trajectory classification and turning movement counts.  We further show that even higher accuracy can be achieved through weak fusion of back-projected detections from multiple cameras.  These results suggeest that traffic should be analyzed on the ground plane, not the image plane.  
\end{abstract}

\section{INTRODUCTION}
Accurate turning movement counts (TMC) are essential data for core intelligent transportation system (ITS) goals, including optimizing traffic signal timings, alleviating congestion, and enhancing road safety. They are also important for the design of efficient road networks, planning pedestrian pathways, and strategizing for future urban expansions. 

Computer vision systems for TMC typically involve detection of vehicles in each image frame, tracking of these detections over frames, and then classification of each image plane trajectory into one of several possible turning movement classes. Accurate classification requires that tracks from different classes are well-separated relative to the dispersion within each class.  While this is typically true on the ground plane, projection to the image plane distorts these trajectories and could potentially degrade this separation. 

This motivates an analysis of whether it is advantageous to perform trajectory classification in the ground plane rather than the image plane.  We perform this analysis here, making the following specific contributions:
\begin{enumerate}
    \item We enhance two public ITS datasets and contribute a third novel dataset, providing ground-truth turn counts, orthographic projections and homographies from the image plane to the ground plane to support ground plane analysis.
    \item We develop and open-source a novel pipeline for automatic ground-plane TMC.
    \item We use our three datasets to evaluate and compare single-camera TMC in the image plane vs the ground plane, finding classification on the ground plane to be consistently more accurate.
    \item We demonstrate how this approach can be extended to multi-camera systems to achieve even more accurate ground plane classification through a simple weak-fusion approach. 
\end{enumerate}

\section{PRIOR WORK}
Historically, turning movement counts have been collected manually or by using inductive loop detectors~\cite{intro2kotzenmacher2005evaluation}.  These methods are reliable but costly.  Manual methods require deployment of personnel every time a cont is required, and inductive loops are susceptible to damage from shifts in the roadbed due, for example, to freeze/thaw cycles. Roadside radar is less expensive and more flexible, but provides limited information about each vehicle~\cite{ZHONG200725}. Since the 1990s, traffic surveillance cameras have become increasingly explored for ITS applications due to reasonable costs and utility for both human and machine interpretation.  

Computer vision approaches for TMC based on traffic camera data typically comprise three stages:  Detection, tracking and classification of trajectories.

\subsection{Detection}
Early traffic analytics methods often relied on background subtraction for detection~\cite{BARCELLOS20151845, 4290146,ShokrolahShirazi2019, XIA2016672}. Modern methods typically use convolutional neural network (CNN) detectors, including two-stage detectors like Faster RCNN~\cite{9150979} and one-stage detectors like CenterNet \cite{9523000}, various versions of YOLO~\cite{9522996, 9522714, 9523072, 9150979},
RetinaNet \cite{9523074}, and NAS-FPN \cite{9150753}. 

\subsection{Tracking}
Tracking methods applied in early traffic analytics systems were typically feature-based, employing particle filters \cite{5457465}, dynamic feature grouping \cite{4587551}, the Kanade-Lucas-Tomasi (KLT) tracker~\cite{7045864}, region-based tracking \cite{4118800, 1640826} and contour-based tracking \cite{1570925}. Modern approaches typically employ online trackers based on variations of the Kalman filter, such as SORT \cite{9523072}, DeepSort \cite{9150961, 9150753}, ByteTrack \cite{ 10421880}, and JDE \cite{9151032}. Further advances have been achieved by training a re-identification feature network \cite{9151055}, implementing Mahalanobis distance smoothing \cite{9150979},  and integration of trajectories from multiple viewpoints \cite{10444381}.
Some traffic analytics systems have  employed  tracking methods like CenterTrack \cite{9523000} and  Tracktor \cite{9150586} that eschew the Kalman model and instead learn to directly predict and associate current detections from the bounding box coordinates in the previous frame.  
\subsection{Classifying trajectories}
Trajectory classification methods can be  categorized into line-based, zone-based, field map-based and prototype-matching, or a combination thereof. Line-based methods~\cite{10444381, 9150586} classify trajectories based on their intersections with predefined virtual lines. Zone-based methods~\cite{9523074, 7045864} classify based on entry and exit zones;
\cite{7045864} incorporates an additional central zone to handle shorter tracks. Field map methods~\cite{9523000,9151032} classify trajectories based upon proximity and completeness field maps derived from manually drawn models of each movement of interest.  Prototype matching techniques define a typical trajectory for each object, either manually~\cite{9522996, 9522714},  semi-automatically using line or zone intersections \cite{9150979, 9150708}, or with unsupervised automatic clustering methods~\cite{9150979, 9523072}. Matching of new trajectories to prototypes is achieved with a range of distance metrics, including cosine distance of trajectory directions, shape-based measures like the Hausdorff distance \cite{9523072}, longest common subsequence (LCSS) \cite{9562804}, nearest neighbor \cite{9150708}, or combinations of these \cite{9522996, 9522714}.
\cite{7384599} explores a cascaded approach, combining zone-based and prototype matching techniques.

Prior work~\cite{9150708, 10444381} has explored back-projection to the ground plane for classification of turning movement counts and integration of multiple camera views, however whether this afford significant advantage over camera-plane classification has not been systematically tested.  

\subsection{Our approach}
We adopt the standard detect-track-classify framework. We select the detector and tracker based upon a comprehensive evaluation on our training data.  
 For detection we employ the deconvolution-based foundation model, InternImage~\cite{InternImage}.  For tracking we use the Kalman filter-based ByteTrack system~\cite{10421880}.
 
 Our main innovation is in trajectory classification.  We use an unsupervised method to identify reliable exemplars for each turning movement class and use these to derive a dense likelihood map for each turning movement class over either the image plane or the ground plane.  Approximating vehicle locations over a trajectory as conditionally independent, we use these maps to estimate the conditional likelihood of novel trajectories tracks.  This method  proves effective for both longer and shorter tracks. 
 
\section{DATASETS}

\begin{figure}[ht]
     \centering
     \begin{subfigure}{\columnwidth}
         \centering
         \includegraphics[width=0.49\textwidth]{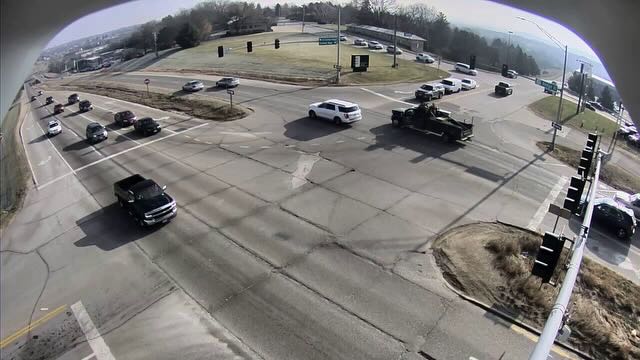}
         \includegraphics[width=0.49\textwidth]{ 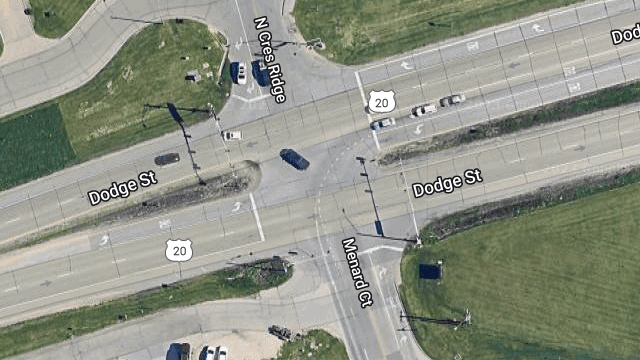}
         \caption{Example from the CityFlow V2 dataset~\cite{AIC22}.}
     \end{subfigure}
\vspace{0.2em}

      \begin{subfigure}{\columnwidth}
         \centering
         \includegraphics[width=0.49\textwidth]{ 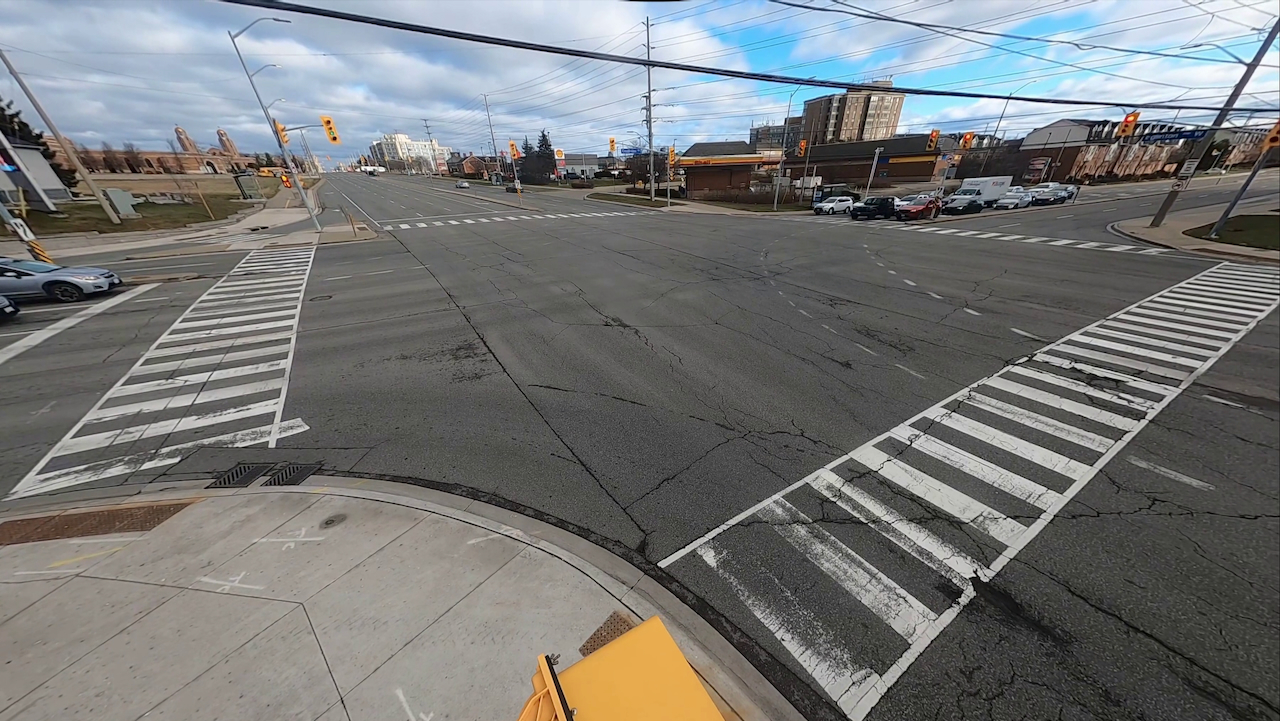}
         \includegraphics[width=0.49\textwidth]{ 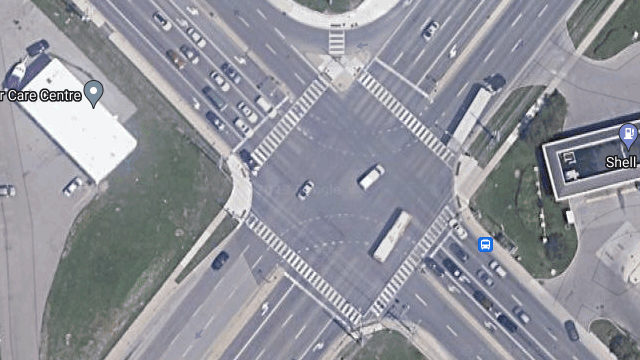}
         \caption{Example from the Trans-Plan dataset.}
     \end{subfigure}
\vspace{0.2em}

     \begin{subfigure}{\columnwidth}
         \centering
         \includegraphics[width=0.49\textwidth]{ 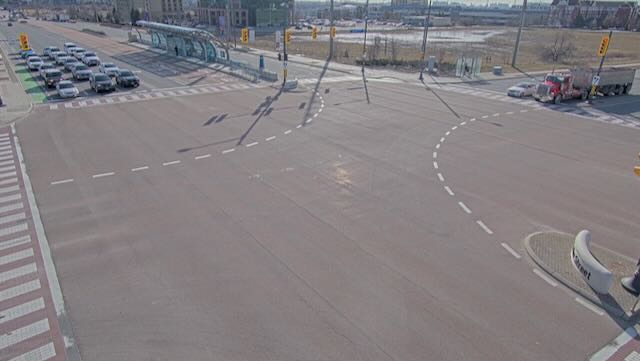}
         \includegraphics[width=0.49\textwidth]{ 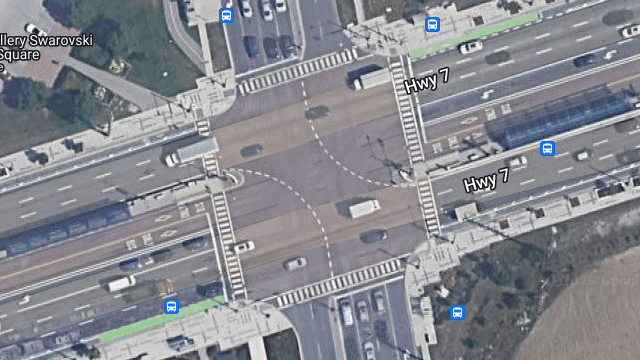}
         \caption{Example from the Region of York dataset.}
     \end{subfigure}
        \caption{Examples from our three datasets.  Left:  Camera views.  Right: Orthophotos from Google Maps.  The Region of York dataset (c) provides camera views from each of the four corners of the intersection;  only one is shown here.}
        \label{fig:datasets_overView}
\end{figure}

Most existing traffic intersection datasets (e.g., \cite{wen2020ua,Naphade20AIC20,AIC21,AIC22,yu2022dair}) provide only one camera view per intersection.  (An exception is the multimodal A9-Intersection dataset~\cite{cress2022A9}, which provides two views per intersection.  However, the field of view of the two cameras is relatively narrow and insufficient to monitor traffic in all directions.)  Here we evaluate camera plane and ground plane TMC methods on the CityFlow V2~\cite{AIC22} dataset (Fig. \ref{fig:datasets_overView}(a)) as a typical representative of these existing datasets.

We  also introduce two new datasets to address issues that cannot be addressed using existing datasets.
In all of these existing datasets, cameras are mounted relatively high in the infrastructure, providing an advantageous view with fewer occlusions \cite{10444381}.  However, for typical traffic surveys, cameras are mounted temporarily at much lower heights (less than 5 metres).  To address this use case, we have worked with our industry partner Trans-Plan to generate a new single-view, low-vantage, wide-angle dataset (Fig. \ref{fig:datasets_overView}(b)). 
To address the
question of whether and to what extent TMC accuracy might depend upon the number of camera views, our second new {\em Region of York dataset} provides four synchronized views of the same intersection (Fig. \ref{fig:datasets_overView}(c)).  

Ground truth turning movement counts were manually collected for all three datasets.  Orthophoto imagery was sourced from Google Maps and used to  establish point correspondences and determine homographies mapping camera planes to planar approximations of the ground terrain.

We divided Trans-Plan and Region of York datasets into training and validation partitions.  Since we do not use any supervised training methods there is no risk of overfitting.  However, we do employ some unsupervised optimization methods that we view as occurring prior to deployment, or perhaps refreshed periodically, so validation data are important to estimating performance at deployment. Since the duration of video for each intersection of the CityFlow V2 dataset is quite limited, we did not partition this dataset, and therefore results reported on this dataset may be slightly optimistic.
We provide more details on each of these three datasets below.
\subsection{CityFlow V2 dataset}
The AI City Challenge 2022 Benchmark Dataset CityFlow-V2~ \cite{CityFlow} consists of traffic surveillance data acquired from a network of 46 cameras mounted in the infrastructure, capturing 880 annotated vehicles. We have selected eight videos (from Cameras 5, 35, 41, 42, 43, 44, 45, and 46), each providing a wide field of view of a different intersection (Fig. \ref{fig:datasets_overView}(a)) at 960p and 10fps, totaling 23m30s of video. 

\subsection{Trans-Plan dataset}
This dataset was acquired by our industry partner Trans-Plan. 
60-minute 4K 60fps videos of four intersections were recorded with a wide-field GoPro camera  strapped to a light pole at a height of roughly 5m. The first half of each video was used for training and the second half for validation.  

\subsection{Region of York dataset}
This dataset consists of synchronized video from four 4K 60fps wide-angle cameras viewing the same intersection from each of its four corners. Two LiDAR systems provide 3D ground truth. Here we make use of 32 20-second segments from the four cameras (i.e., a total of 128 videos), partitioned into 25 training segments and 7 validation segments, capturing a wide range of weather and illumination scenarios. A paper introducing this dataset is currently under review.

\section{METHODS}

Our TMC approach consists of three phases:  A) Geometric calibration, B) Unsupervisd learning and 3) Inference.

\subsection{Geometric calibration}
\begin{figure}[htbp]
     \centering
     \begin{subfigure}{\columnwidth}
         \centering         \includegraphics[width=0.49\textwidth]{ 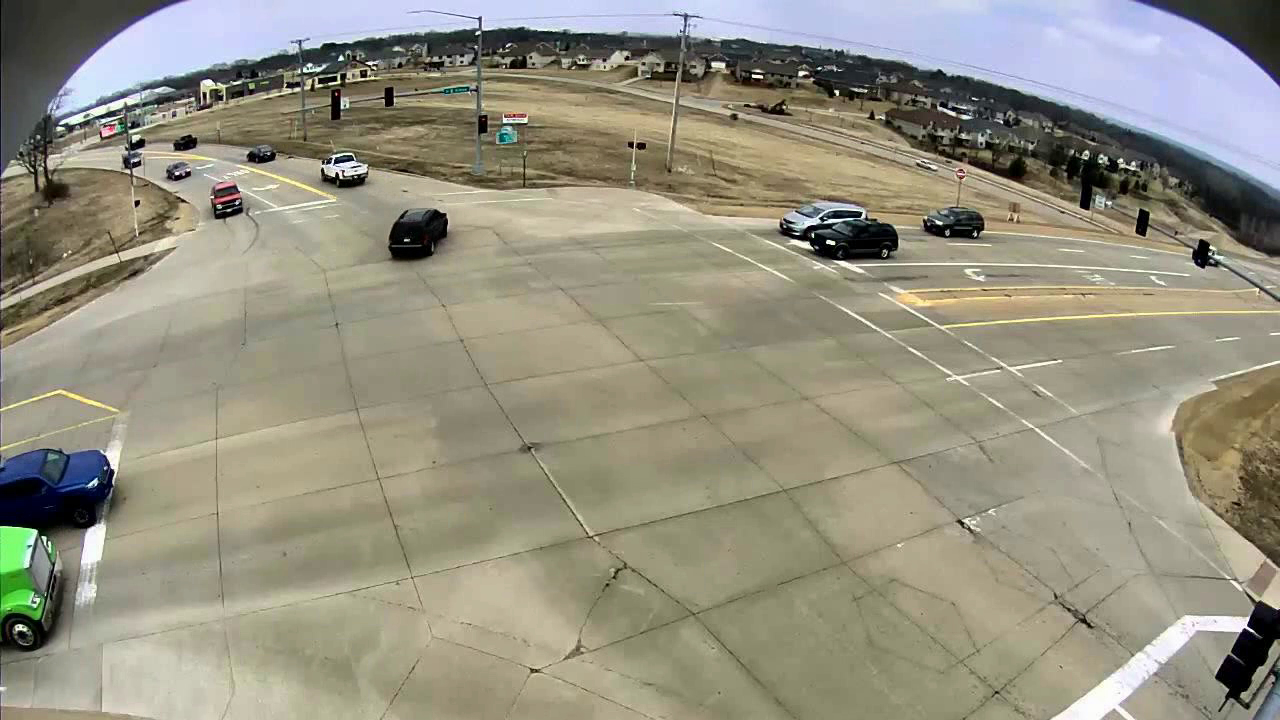}
        \includegraphics[width=0.49\textwidth]{ 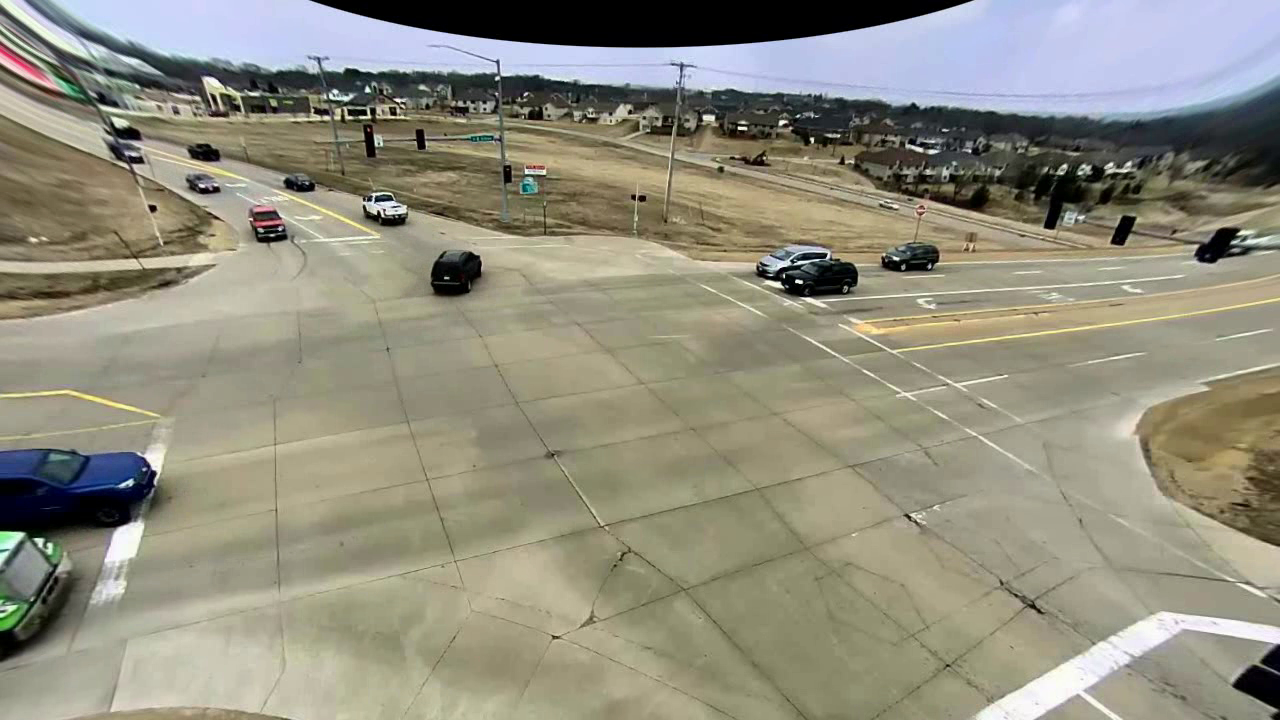}
        \caption{Intrinsic calibration.  Left: Original image.  Right: With nonlinear distortion removed.  Example  taken from CityFlow V2.}
        \label{fig:distortion}
     \end{subfigure}
        \vspace{0.5em}
        
       \begin{subfigure}{\columnwidth}
         \centering        \includegraphics[width=0.49\textwidth]{ 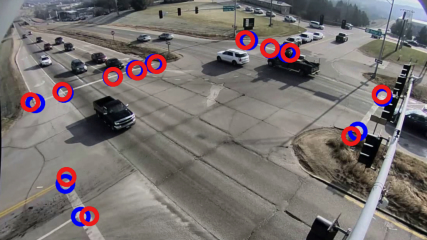}
        \includegraphics[width=0.49\textwidth]{ 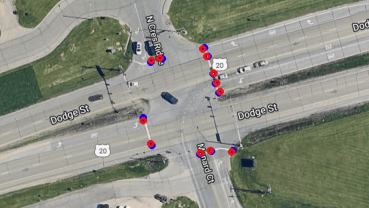}
        \caption{Extrinsic calibration.  Corresponding keypoints (red) are selected in  a camera frame (left) and orthophoto (right) and used to estimate a homography mapping between camera and ground planes. Reprojected keypoints are shown in blue. 
 Example  taken from CityFlow V2.} \label{fig:homographyGUIPairPoints}
 \end{subfigure}
        \vspace{0.5em}
        
     \begin{subfigure}{\columnwidth}
         \centering
        \includegraphics[width=0.49\textwidth]{ 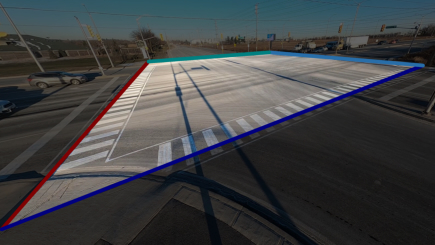}
       \includegraphics[width=0.49\textwidth]{ 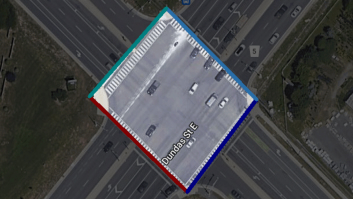}
                \caption{Region of interest (ROI) defined in camera plane (left) and orthophoto (right).  Example taken from the Trans-Plan dataset.}
        \label{fig:ROI}
      \end{subfigure}
      \caption{Three steps of geometric calibration:  (a) Intrinsic  calibration; (b) Extrinsic calibration; (c) Region of interest annotation.}
\label{fig:calibration}
\end{figure}

Calibration is performed once for each deployment site, and consists of three steps (Fig. \ref{fig:calibration}):  1) Intrinsic camera calibration, 2) Extrinsic calibration and 3) Region of interest annotation.

\subsubsection{Intrinsic camera calibration}
Intrinsic parameters of the CityFlow cameras are provided as part of the dataset.  We calibrated the GoPro camera used for the Trans-Plan dataset and the Axis cameras used for the Region of York dataset using standard methods in our lab.  We use these intrinsic parameters to remove radial distortion and thus more accurately approximate a planar projection.

\subsubsection{Extrinsic camera calibration}
We do not attempt a full estimation of extrinsic parameters but instead estimate for each site and each camera a homography mapping from the camera plane to a planar approximation of the ground plane.  To do this, we obtain an orthophoto from Google Maps and manually identify corresponding keypoints in the orthophoto and an example frame from each camera at the site.  Table \ref{tab:Homography_errors} shows the mean number of keypoints selected per image and resulting reprojection errors.

  \begin{table}[ht]
    \centering
    \begin{tabular}{@{}lccccc@{}}
         \toprule
         \textbf{Dataset} & \textbf{\# Points}& 
          \textbf{Camera res.}&
          \multicolumn{2}{c}{\textbf{Reprojection error}}\\ 
&&
{\textbf{(pixels)}}
&
\multicolumn{2}{c}{\textbf{(pixels)}}\\ 
&&& \textbf{Camera} & \textbf{Orthophoto} \\
       \midrule
         TransPlan & 22 & 3596 x 2158 & 18.2 & 4.3 \\
         \midrule
         CityFlow & 13 & 1280 × 960 & 16.7 & 4.2 \\
         \midrule
         Region of York & 18 & 4840 × 2726 & 15.8 & 2.8 \\
         \bottomrule
    \end{tabular}
    \captionof{table}{Average number of keypoints and resulting reprojection errors for estimated homographies on our three datasets. All orthophotos are 640 $\times$ 640 pixels.}
    \label{tab:Homography_errors}
  \end{table}
  
\subsubsection{Region of interest annotation}
For each intersection, we define a region of interest (ROI) in each orthophoto defined by the four corners of the intersection. We also identify the number of lanes for each  turning movement class.

\subsection{Unsupervised learning}
\begin{figure*}[ht]
     \centering
       \begin{subfigure}{1.0\textwidth}
         \centering
         \includegraphics[width=\textwidth]{ 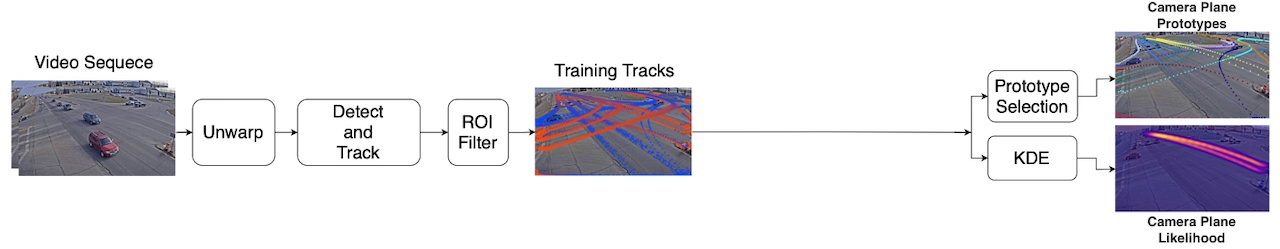}
         \caption{Camera plane.}
     \end{subfigure}
       \begin{subfigure}{1.0\textwidth}
         \centering
         \includegraphics[width=\textwidth]{ 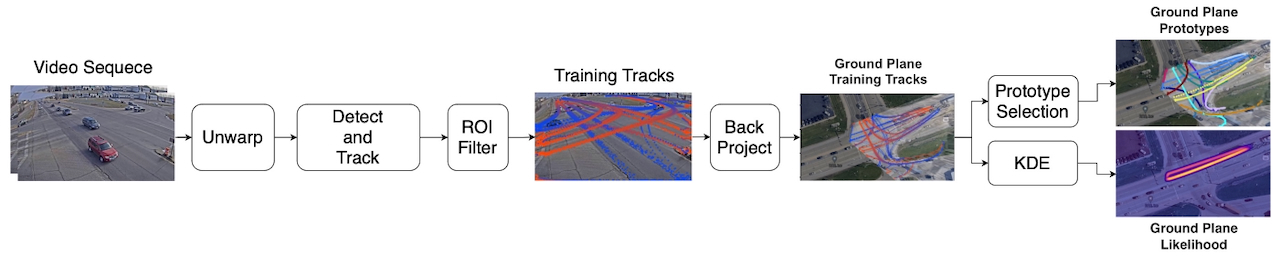}
         \caption{Ground plane.}
     \end{subfigure}
        \caption{Unsupervised learning pipelines for (a) camera plane and (b) ground plane systems.}
        \label{fig:Opt_Pipeline}
\end{figure*}

Having geometrically calibrated the system, our goal is to use observations of vehicle dynamics in the training videos to optimize TMC in unsupervised fashion.  This process consists of:  1) Object detection, 2) Tracking,  3) Back-projection (for ground plane method) and 4) Turning movement modeling.  (Fig. \ref{fig:Opt_Pipeline}).

\subsubsection{Object Detection and Tracking}

In a recent comprehensive comparison of object detection and tracking on the Region of York dataset (under review for another conference), we found that a combination of the InternImage \cite{InternImage} detector with the ByteTrack tracker \cite{bytetrack} consistently outperformed other systems so we use this combination here.

A track is defined as a sequence of positions in the image or on the ground plane.  Due to vehicle accelerations, samples taken uniformly over time are typically not uniform over space, which complicates analysis and classification.  To address this, tracks are first resampled uniformly in space using linear interpolation at 5 pixel resolution in the camera plane and 20cm resolution on the ground plane. 

\subsubsection{Back-projection to ground plane}
For the ground plane version of our TMC system, objects detected in the image plane are back-projected to our planar ground plane model.  Specifically, we back-project the midpoint of the bottom segment of the estimated bounding box.  Note that this is an imperfect method for geo-localization as this point will generally be displaced relative to the gravity projection of the vehicle centroid, and this displacement will depend upon the pose of the vehicle.  Future research could explore using 3D object detectors or other strategies for improving the accuracy of back-projection.

\subsubsection{Turning movement modeling}
Our classification strategy is to use training data to automatically select {\em training} tracks that are representative of each different turning movement.  At inference, a new track is then compared to these training data to determine its movement class.

To define training tracks we use a fully automatic version of a semi-automatic approach that has been used in previous systems~\cite{9150979}. Specifically, we define a training track as a track that enters the ROI through one of its four defining line segments and exits through another.  The class of the track is then determined by the entrance and exit segment, which defines $4\times3 = 12$ unique turning movement classes.

Having defined the training tracks for each turning movement class, we explore two different strategies for summarizing the geometry of these tracks for each class to support efficient inference.

\begin{figure}[htbp]
     \centering
     \begin{subfigure}{0.23\textwidth}
         \centering
         \includegraphics[width=\textwidth]{ 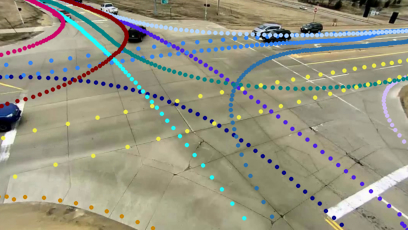}
     \end{subfigure}
     \vspace{3pt}
     \begin{subfigure}{0.23\textwidth}
         \centering
         \includegraphics[width=\textwidth]{ 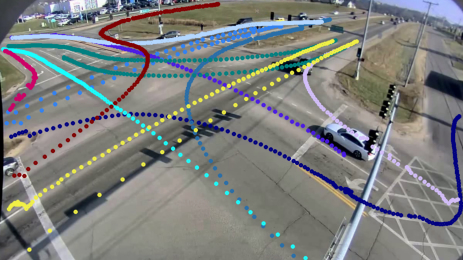}
     \end{subfigure}
     \begin{subfigure}{0.23\textwidth}
         \centering
         \includegraphics[width=\textwidth]{ 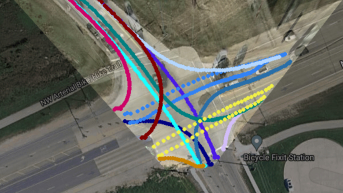}
     \end{subfigure}
      \begin{subfigure}{0.23\textwidth}
         \centering
         \includegraphics[width=\textwidth]{ 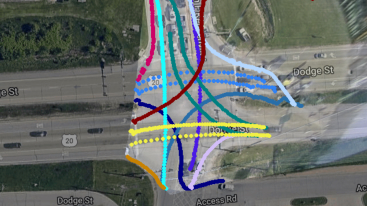}
     \end{subfigure}
     
        \caption{Example representative prototype tracks selected using our unsupervised clustering method, for two CityFlow V2 sites.  One track is selected for each lane within each turning movement class.  Each class is represented by a different colour.  Top:  image plane.  Bottom:  orthophoto.  }
        \label{fig:prototypes}
\end{figure}

{\bf Prototype model}
Our first strategy follows a standard approach~\cite{9150753, 7384599, 9523072} in which each lane of each turning movement is represented by a single representive prototype track.  One novelty of our approach is that we select these prototypes in an unsupervised way.  Specifically, we first cluster the training tracks within each movement class using the k-means++ algorithm \cite{kmeansplusplus}, where the number of clusters $k$ is equal to the number of lanes within the movement class, annotated at the Geometric Calibration stage. We then select from each cluster the training track that minimizes the average distance to other training tracks within the cluster, using the contour mapping measure (CMM) as a distance metric \cite{cmm}. This process yields lane-specific prototypes for each turning movement of interest. Prototype tracks computed for two example intersections are shown in Fig.~\ref{fig:prototypes}.

{\bf Probability model}
In our second strategy, we form a probabilistic model over the distribution of trajectories within each turning movement class.  Here we employ Gaussian kernel density estimation (KDE) and do not rely upon explicit clustering of lanes with each movement class.

Our goal is to form a Gaussian KDE model of the conditional likelihood $p\left(T_i|M_j\right)$ of a track $T_i$ given it originates from movement class $M_j$.   

Each track $T_i$ consists of a sequence of $n_i$ 2D positions: 

\begin{equation}
    T_i = \left\{{\mathbf x}_{i1},\ldots,{\mathbf x}_{in_i}\right\}. 
\end{equation} 

We model the likelihood of each of these positions as conditionally independent, so that the conditional likelihood each track $T_i$ is given by 

\begin{equation}
p\left(T_i|M_j\right)=\prod_{k=1\ldots n_i}p\left(\mathbf{x}_{ik}|M_j\right).
\end{equation} 

Note that this approximation discards information about the temporal order of the positions, which could limit performance.  This is one area for potential future research.

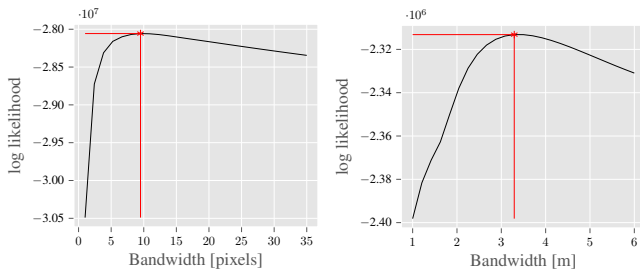
\begin{figure}[ht]
\centering
\resizebox{0.49\linewidth}{!}{
\begin{tikzpicture}

\definecolor{dimgray85}{RGB}{85,85,85}
\definecolor{gainsboro229}{RGB}{229,229,229}

\begin{axis}[
axis background/.style={fill=gainsboro229},
axis line style={white},
tick align=outside,
tick pos=left,
x grid style={white},
xlabel=\textcolor{dimgray85}{Bandwidth [pixels]},
xlabel style={font=\Large},
ylabel style={font=\Large},
xmajorgrids,
xmin=-0.7, xmax=36.7,
xtick style={color=dimgray85},
y grid style={white},
ylabel=\textcolor{dimgray85}{log likelihood},
ymajorgrids,
ymin=-30611899.6475794, ymax=-27935498.4864569,
ytick style={color=dimgray85},
ytick={-31000000,-30500000,-30000000,-29500000,-29000000,-28500000,-28000000,-27500000},
yticklabels={
  \ensuremath{-}3.10,
  \ensuremath{-}3.05,
  \ensuremath{-}3.00,
  \ensuremath{-}2.95,
  \ensuremath{-}2.90,
  \ensuremath{-}2.85,
  \ensuremath{-}2.80,
  \ensuremath{-}2.75
}
]
\addplot [semithick, black]
table {%
1 -30490245.0493465
2.41666666666667 -28724288.3907219
3.83333333333333 -28311164.5790498
5.25 -28160146.6611166
6.66666666666667 -28099758.5369677
8.08333333333333 -28068740.2467791
9.5 -28057153.0846897
10.9166666666667 -28060157.8365166
12.3333333333333 -28070594.303576
13.75 -28086031.4021573
15.1666666666667 -28103884.0669006
16.5833333333333 -28122095.4219241
18 -28139631.7323367
19.4166666666667 -28157170.4664837
20.8333333333333 -28174869.8575587
22.25 -28192676.3859939
23.6666666666667 -28210383.3747341
25.0833333333333 -28227948.4623139
26.5 -28245331.3571859
27.9166666666667 -28262504.5602476
29.3333333333333 -28279463.7482305
30.75 -28296216.3184844
32.1666666666667 -28312774.285551
33.5833333333333 -28329150.414345
35 -28345356.3221897
};
\addplot [semithick, red, mark=asterisk, mark size=3, mark options={solid}]
table {%
9.5 -28057153.0846897
};
\addplot [semithick, red]
table {%
9.5 -30490245.0493465
9.5 -28057153.0846897
};
\addplot [semithick, red]
table {%
1 -28057153.0846897
9.5 -28057153.0846897
};
\end{axis}

\end{tikzpicture}}
\resizebox{0.49\linewidth}{!}{
\begin{tikzpicture}

\definecolor{dimgray85}{RGB}{85,85,85}
\definecolor{gainsboro229}{RGB}{229,229,229}

\begin{axis}[
axis background/.style={fill=gainsboro229},
axis line style={white},
tick align=outside,
tick pos=left,
x grid style={white},
xlabel=\textcolor{dimgray85}{Bandwidth [m]},
xlabel style={font=\Large},
ylabel style={font=\Large},
xmajorgrids,
xmin=0.75, xmax=6.25,
xtick style={color=dimgray85},
y grid style={white},
ylabel=\textcolor{dimgray85}{log likelihood},
ymajorgrids,
ymin=-2402407.49288164, ymax=-2308926.38705869,
ytick style={color=dimgray85},
ytick={-2420000,-2400000,-2380000,-2360000,-2340000,-2320000,-2300000},
yticklabels={
  \ensuremath{-}2.42,
  \ensuremath{-}2.40,
  \ensuremath{-}2.38,
  \ensuremath{-}2.36,
  \ensuremath{-}2.34,
  \ensuremath{-}2.32,
  \ensuremath{-}2.30
}
]
\addplot [semithick, black]
table {%
1 -2398158.35170787
1.20833333333333 -2381600.39699554
1.41666666666667 -2371190.92870603
1.625 -2362724.68905332
1.83333333333333 -2350101.75228651
2.04166666666667 -2337940.22104304
2.25 -2328685.65335015
2.45833333333333 -2322264.54082262
2.66666666666667 -2318018.38468741
2.875 -2315344.08593982
3.08333333333333 -2313826.06071203
3.29166666666667 -2313175.52823246
3.5 -2313191.77861428
3.70833333333333 -2313718.41644957
3.91666666666667 -2314635.10488611
4.125 -2315844.71268844
4.33333333333333 -2317270.91385176
4.54166666666667 -2318853.19401429
4.75 -2320543.11207084
4.95833333333333 -2322298.9804735
5.16666666666667 -2324077.30965546
5.375 -2325855.81761208
5.58333333333333 -2327615.51584886
5.79166666666667 -2329307.46169123
6 -2330940.9991671
};
\addplot [semithick, red, mark=asterisk, mark size=3, mark options={solid}]
table {%
3.29166666666667 -2313175.52823246
};
\addplot [semithick, red]
table {%
3.29166666666667 -2398158.35170787
3.29166666666667 -2313175.52823246
};
\addplot [semithick, red]
table {%
1 -2313175.52823246
3.29166666666667 -2313175.52823246
};
\end{axis}

\end{tikzpicture}}
\caption{Average log likelihood of training tracks for the second half of the training partition for one Trans-Plan site based on KDE model derived from the first half of the training partition.  The likelihood peaks at 9.7 pixels in the image plane (left) and 3.3 metres in the ground plane (right).}
\label{fig:KDEBW}
\end{figure}

We use KDE to estimate the probability density for vehicle positions at pixel resolution in the camera plane and 22cm resolution in the ground plane.
We optimize the bandwidth of the Gaussian kernel using the training partition of one of the four Trans-Plan sites.  We first partitioned this 30-minute segment into two 15-minute segments. We then swept over a range of bandwidths, forming the conditional likelihood models based upon the first 15-minute partition and evaluating the conditional likelihood of all training tracks of the second partition.  Fig. \ref{fig:KDEBW}  shows the average conditional log likelihood over tracks from the second partition as we sweep over bandwidth.  We found the optimal bandwidth to be 9.7 pixels in the image domain and 3.36 metres in the ground plane.

\begin{figure*}[ht]
     \centering
     \begin{subfigure}{0.16\textwidth}
         \centering
         \includegraphics[width=\textwidth]{ 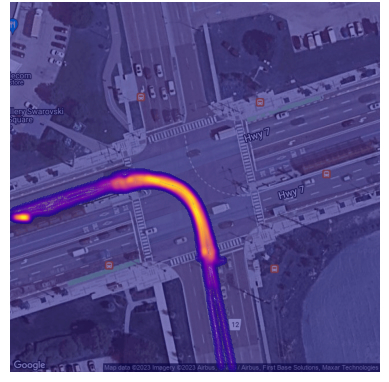}
     \end{subfigure}
     \begin{subfigure}{0.16\textwidth}
         \centering
         \includegraphics[width=\textwidth]{ 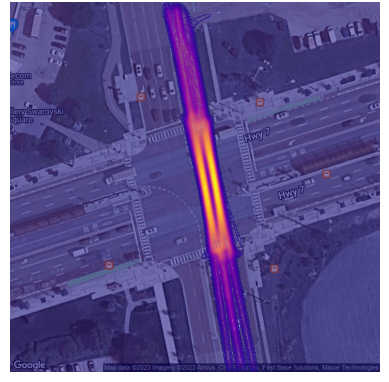}
     \end{subfigure}
     \begin{subfigure}{0.16\textwidth}
         \centering
         \includegraphics[width=\textwidth]{ 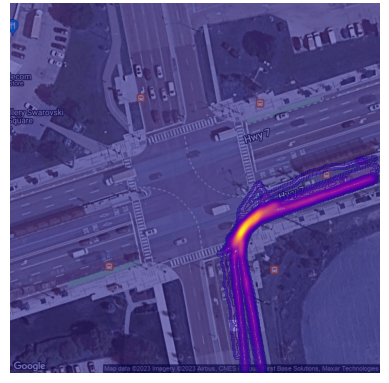}
     \end{subfigure}
      \begin{subfigure}{0.16\textwidth}
         \centering
         \includegraphics[width=\textwidth]{ 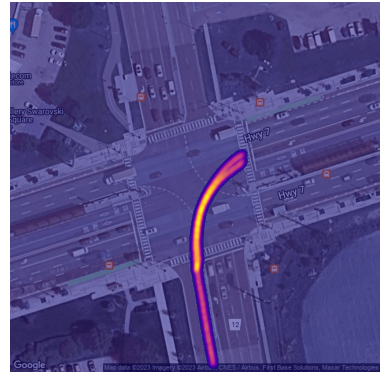}
     \end{subfigure}
       \begin{subfigure}{0.16\textwidth}
         \centering
         \includegraphics[width=\textwidth]{ 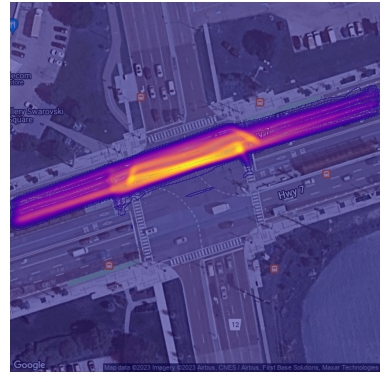}
     \end{subfigure}
    \begin{subfigure}{0.16\textwidth}
         \centering
         \includegraphics[width=\textwidth]{ 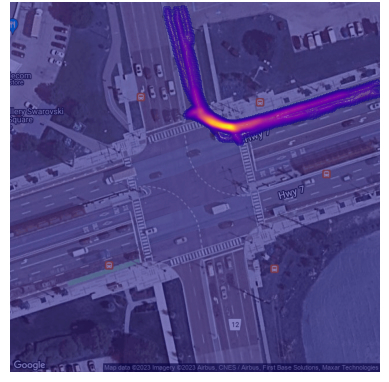}
     \end{subfigure}
     \begin{subfigure}{0.16\textwidth}
         \centering
         \includegraphics[width=\textwidth]{ 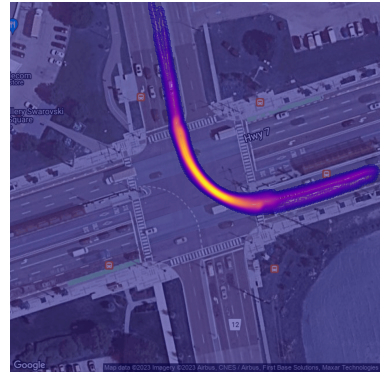}
     \end{subfigure}
      \begin{subfigure}{0.16\textwidth}
         \centering
         \includegraphics[width=\textwidth]{ 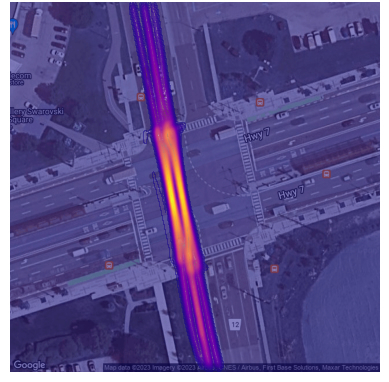}
     \end{subfigure}
           \begin{subfigure}{0.16\textwidth}
         \centering
         \includegraphics[width=\textwidth]{ 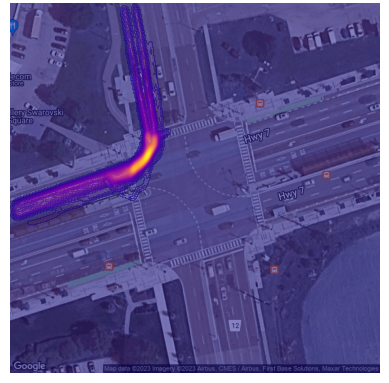}
     \end{subfigure}
           \begin{subfigure}{0.16\textwidth}
         \centering
         \includegraphics[width=\textwidth]{ 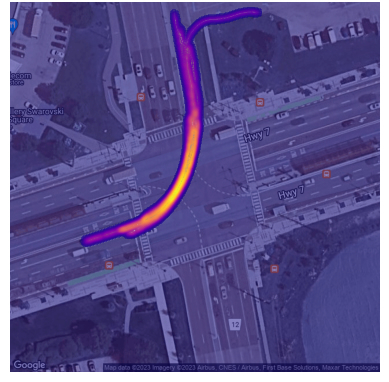}
     \end{subfigure}
           \begin{subfigure}{0.16\textwidth}
         \centering
         \includegraphics[width=\textwidth]{ 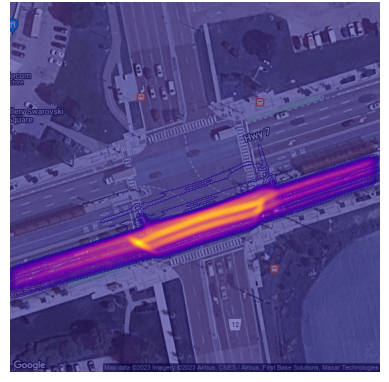}
     \end{subfigure}
           \begin{subfigure}{0.16\textwidth}
         \centering
         \includegraphics[width=\textwidth]{ 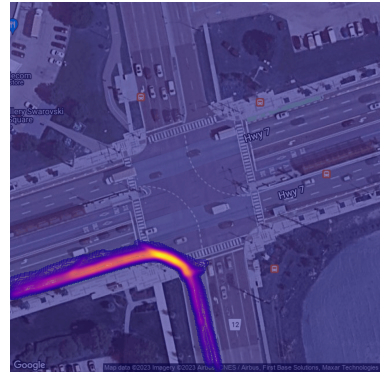}
     \end{subfigure}
        \caption{Optimized likelihood models in the ground plane for the 12 movement classes at the Region of York intersection.}
        \label{fig:KDEDensities}
\end{figure*}

Fig. \ref{fig:KDEDensities} shows optimized likelihood models in the ground plane for the 12 turning movement classes for the Region of York intersection.  

\subsection{Inference}
\begin{figure*}[ht]
     \centering
       \begin{subfigure}{1.0\textwidth}
         \centering
         \includegraphics[width=\textwidth]{ 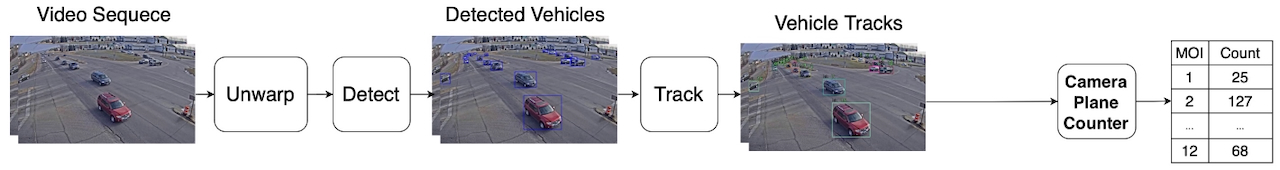}
         \caption{Camera plane inference pipeline.}
     \end{subfigure}
     \vspace{0.5em}
     
       \begin{subfigure}{1.0\textwidth}
         \centering
         \includegraphics[width=\textwidth]{ 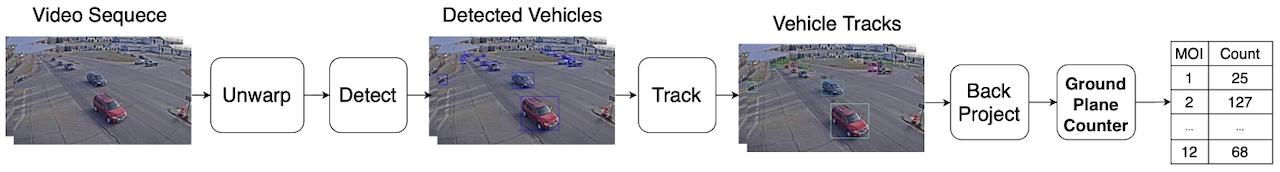}
         \caption{Ground plane inference pipeline}
     \end{subfigure}
        \caption{Inference pipelines for (a) camera plane and (b) ground plane methods.}
        \label{fig:Infer_Pipeline}
\end{figure*}

Our inference pipeline mirrors our optimization pipeline and consists of (Fig. \ref{fig:Infer_Pipeline}): 1) Object detection, 2) Tracking,  3) Back-projection of tracks to the ground plane (for our ground plane system) and 4) Turning movement classification.  The first three steps have already been described above in the context of our unsupervised learning phase;  here we focus on turning movement classification.

\subsubsection{Turning movement classification.}
We evaluate four distinct methods for turning movement classification:

{\bf Entry-Exit (EE) method}
The entry-exit method does not use learned turning movement models but simply classifies an inference track based on its geometry relative  to the four line segments bounding the ROI.  In particular, inference tracks that cross two different segments are classified exactly as training tracks are classified (see above).  If the inference track starts or ends within the ROI, the closest ROI line segment is assigned as the entry or exit segment.  Note that we do not evaluate this method for the Region of York dataset where, due to limitations in the camera field of view, many tracks do not extend to the edges of the ROI.

{\bf Direction (DIR) method}
Like the entry-exit method, the direction method only uses the start and end points of the inference track.  However, unlike the entry-exit method, the direction method makes use of the prototype tracks selected during the unsupervised learning stage of our system.  In particular, the direction method compares the overall direction of the inference track to the overall directions of the prototype tracks, assigning the turning movement class of the prototype with direction that is most similar to the direction of the inference track.

The direction of each prototype track is defined by the unit vector pointing from its entry point to its exit point.  For inference tracks that extend across the ROI, direction is defined in the same way.  For inference tracks that start and/or end within the ROI, the start and/or endpoint are employed to define the direction.  Direction similarity is defined by the dot product of the unit vectors (cosine distance).

{\bf Voting (VOTE) method}
While the entry-exit and direction methods use only two points on the inference track, our remaining two methods examine the entire track. The voting method marches over the points of the inference track, for each point casting a vote for the turning movement class with the closest prototype to that point.  The inference track is then assigned to the  movement class with the most votes.

{\bf Maximum Likelihood (ML) method}
While the direction and voting methods make use of  prototype tracks, our maximum likelihood method relies upon a learned likelihood model, computing the conditional likelihood of each inference track for each turning movement class and assigning the class with the highest likelihood.  While generally the prior distribution over movement classes is far from uniform, we refrain from applying a prior here, partly because it is not clear how stable this prior would be over time of day.  This remains an interesting issue for future research.

\section{EXPERIMENTS AND RESULTS}

\subsection{Single-view turning movement counts}
Table \ref{tab:turn-counts-2D-vs-3D} compares turn count error rates on the CityFlow dataset and on the validation partitions of our TransPlan and Region of York datasets. We report mean absolute error over all turning movement classes. We evaluate all four of our turning movement count classification methods in both camera plane and ground plane domains.

\begin{table}[ht]
\centering
    \begin{tabular}{@{}lccccc@{}}
         \toprule
         \textbf{Dataset} & \textbf{Domain} & \textbf{EE}& \textbf{DIR}& \textbf{VOTE}& \textbf{ML} \\
         \midrule
         CityFlow & Camera & 16.0\% & 6.6\% & 7.8\%   & 5.1\% \\
         \cmidrule{2-6}
                 & Ground & 14.0\% & 5.4\% & 7.9\%  &  \textbf{4.0\%} \\
         \midrule
         TransPlan & Camera & 17.7\% & 12.6\% & 14.2\% &  11.0\%\\
         \cmidrule{2-6}
                 & Ground & 14.3\% & 10.5\% & 13.8\% & \textbf{9.9\%} \\
         \midrule
          Region of York & Camera & -------- & 20.0\% & 38.2\% & 19.5\% \\
         \cmidrule{2-6}
                 & Ground & -------- &  14.9\% & 38.2\% & \textbf{13.8\%} \\
         \bottomrule
    \end{tabular}
    \captionof{table}{TMC error rates across various four classification methods on both camera and ground planes.  We evaluate on the training partition of the CityFlow dataset and  on the validation partitions of Trans-Plan and Region of York  datasets.}
    \label{tab:turn-counts-2D-vs-3D}
  \end{table}
  
  We find that the direction method performs substantially better than the entry-exit method, presumably due to the finer-grained information about trajectory direction provided by the prototypes.  Interestingly, despite using the entire trajectory, our voting method performs worse than our direction method.  This suggests that the endpoints of the tracks may in fact be more diagnostic of the turning movement class than interior points and should be weighted more heavily.  

We find that our maximum likelihood (ML) method performs best of all four methods evaluated, on all three datasets.  This indicates that, while a voting strategy is insufficient. it can be beneficial to consider the whole track.  We note that by summing the log likelihood over all points on the inference track, the ML method allows points that are highly deviant from the conditional model to have more than an equal vote and ultimately overrule inliers.  

With only two exceptions we see a consistent improvement when we classify tracks on the ground plane rather than the image plane.  This makes sense, as projection to the image distorts and potentially entangles the trajectories, and renders stationarity assumptions implicit in distance measurements suspect.
Ultimately our best performance is achieved using our maximum likelihood method on the ground plane.

Overall, lowest error rates are achieved on the CityFlow dataset.  We believe this is due to the wide field of view, which provides sightlines beyond the boundaries of the ROI, and the relatively high vantage, which minimizes occlusions.

The higher error rates seen in the TransPlan data are likely due to the lower height of the camera, which leads to more occlusions, squeezes the projection of different turning movement classes more closely together due to projection, and increases the error in back-projection due to the high obliquity of the ground plane.  

The highest errors are encountered with the Region of York dataset.  These arise because of the limited field of view (86 deg horizontal) of the cameras, which means that none of the cameras see the complete ROI.

  \subsection{Multi-view turning movement counts}
  Highest error rates were obtained with our Region of York dataset due to the limited camera field of view.  Fortunately, that field site provides data from four cameras mounted on each corner of the intersection, which raises the possibility of improving TMC performance by fusing these data.

  While a long-term objective is to fuse the data in the ground plane at the object detection stage, here we evaluate a much simpler ``weak fusion" approach in which each camera is assigned responsibility for a disjoint subset of turning movement classes.

  To form these assignments, we evaluate count error rates on the training partition for each of the cameras on each of the 12 turning movement classes, and assign each movement class to the camera generating the lowest error (Table \ref{tab:turn-counts-moi-camera}).  In the case of a tie, assignments are made to balance the load across cameras. This resulted in assigning between 2-4 classes to each camera.

\begin{table}[ht]
\centering
    \begin{tabular}{@{}lrrrr@{}}
         \toprule
         \textbf{TMC}&  \textbf{Cam1} & \textbf{Cam2} &\textbf{Cam3} &\textbf{Cam4}\\
         
        \midrule
         1 & 9.5\%   & 10.5\% & \textbf{4.7\%}  & 33.3\% \\
         \midrule
         2 & 5.0\%      & 3.7\%  & \textbf{2.5\%}   & 5.0\%     \\ 	
         \midrule
         3 & \textbf{50.0\%}     & 50.0\%    & 88.8\% & 77.7\% \\ 	
         \midrule
         4 & \textbf{0.0\%}      & 0.0\%     & 100.0\%     & 400.0\% \\
         \midrule
         5 & 4.2\%   & \textbf{3.1\%}  & 9.5\%  & 24.4\% \\
         \midrule
         6 & 40.0\%     & 40.0\%    & 70.0\%    & \textbf{20.0\%}    \\
         \midrule
         7 & 18.5\%  & 7.4\%  & 18.5\% & \textbf{0.0\%}     \\  
         \midrule
         8 & \textbf{6.4\%}   & 7.6\%  & 8.9\%  & 7.6\%  \\
         \midrule
         9 & 20.0\%     & 28.5\% & 50.0\%    & \textbf{14.2\%} \\
         \midrule
         10 & 66.6\% & \textbf{33.3\%} & 66.6\% & 100.0\%   \\
         \midrule
         11 & 10.7\% & 5.7\%  & 6.0\%  & \textbf{5.7\%}  \\
         \midrule
         12 & 53.8\% & 63.6\% & \textbf{38.4\%} & 76.9\% \\
         \bottomrule
         
    \end{tabular}
    \captionof{table}{Error rates for each camera on each turning movement class, for the Region of York training partition.  {\bf Boldface}: assignment of movement classes to cameras.}
    \label{tab:turn-counts-moi-camera}
  \end{table}

Table \ref{tab:turn-counts-MC-integrated} shows how even weak fusion of the four cameras can substantially reduce error rate as well as bias.  Note that the single-view system tends to over-count, presumably due to fragmentation of tracks caused by occlusions from other vehicles and infrastructure.

  \begin{table}[ht]
\centering
    \begin{tabular}{@{}lcrr@{}}
         \toprule
         \textbf{Mode} &\textbf{Error rate} &\textbf{Bias}\\

        \midrule
         Single     & 13.8\%  & 5.2\%  \\ 	
         \midrule
         Multi & 7.3\%   & 0.1\% \\

         \bottomrule
    \end{tabular}
    \captionof{table}{TMC performance (mean absolute error and mean bias over turning movement classes) of single- and multi-camera (weak fusion) systems on the Region of York validation partition.  }
    \label{tab:turn-counts-MC-integrated}
  \end{table}
  
\section{CONCLUSION \& FUTURE WORK}
Our study has highlighted advantages of ground plane reasoning over camera plane reasoning for accurate traffic analysis at intersections. We also found advantages of unsupervised probabilistic modeling over heuristic classification methods, and demonstrated the advantages of multi-view ove single-view surveillance.

There are many opportunities to extend and improve our approach.  We currently use homographies to map from camera plane to ground coordinates.  However, many intersections are not perfectly flat.  Future work will employ digital terrain models for more accurate back-projection.  

Camera sway also affects the accuracy of back-projection, particularly for the Trans-Plan dataset due to the less rigid pole employed.  This can can potentially be addressed using image stabilization techniques.

We currently back-project 2D detections using the midpoint of the bottom edge of the 2D bounding box.  This method generates offsets from the gravity projection of the centre of mass which are pose-dependent.  Future work could explore the use of 3D detection methods.  Another potential direction is to perform 2D instance segmentation and develop methods for reasoning about the shape of the 2D segmentations to reduce ground plane localization error.

We elected to use a uniform prior over turning movement classes, even though the distribution is typically non-uniform.  Future work could look at benefits of learning and applying an empirical prior, and the possibility of adapting that prior based upon time of day and other potential factors.

Single-view methods for traffic analytics are inherently limited due to field of view constraints and occlusions.  We used a very simple ``weak fusion" approach to fuse data from multiple cameras, but this effectively ignores 75\% of the data.  A better approach is to back-project earlier in the pipeline, at the detection stage, and to fuse detections in 3D coordinates to generate more complete tracks.

\bibliographystyle{plain}
\bibliography{refs}

@inproceedings{bytetrack,
  title={Bytetrack: Multi-object tracking by associating every detection box},
  author={Zhang, Yifu and Sun, Peize and Jiang, Yi and Yu, Dongdong and Weng, Fucheng and Yuan, Zehuan and Luo, Ping and Liu, Wenyu and Wang, Xinggang},
  booktitle={European Conference on Computer Vision},
  pages={1--21},
  year={2022},
  organization={Springer}
}

@InProceedings{Naphade20AIC20,
author = {Milind Naphade and Shuo Wang and David C. Anastasiu and Zheng Tang and Ming-Ching Chang and Xiaodong Yang and Liang Zheng and Anuj Sharma and Rama Chellappa and Pranamesh Chakraborty},
title = "The 4th AI City Challenge",
booktitle = "The IEEE Conference on Computer Vision and Pattern Recognition (CVPR) Workshops",
month = "June",
year = {2020},
pages = "2665–2674"
}

@INPROCEEDINGS{cmm,
  author={Movahedi, Vida and Elder, James H.},
  booktitle="2010 IEEE Computer Society Conference on Computer Vision and Pattern Recognition - Workshops",
  title="Design and perceptual validation of performance measures for salient object segmentation",
  year={2010},
  pages={49-56},
  doi={10.1109/CVPRW.2010.5543739}
}

@InProceedings{AIC21,
author = {Milind Naphade and Shuo Wang and David C. Anastasiu and Zheng Tang and Ming-Ching Chang and Xiaodong Yang and Yue Yao and Liang Zheng and Pranamesh Chakraborty and Christian E. Lopez and Anuj Sharma and Qi Feng and Vitaly Ablavsky and Stan Sclaroff},
title = "The 5th AI City Challenge",
booktitle = "The IEEE Conference on Computer Vision and Pattern Recognition (CVPR) Workshops",
month = "June",
year = {2021},
}

@inproceedings{CityFlow,  
author = {Zheng Tang and Milind Naphade and Ming-Yu Liu and Xiaodong Yang and Stan Birchfield and Shuo Wang and Ratnesh Kumar and David Anastasiu and Jenq-Neng Hwang},  
title = "CityFlow: A city-scale benchmark for multi-target multi-camera vehicle tracking and re-identification",  
booktitle = "Proceedings of the IEEE/CVF Conference on Computer Vision and Pattern Recognition (CVPR)",  
pages = {8797--8806},  
address = "Long Beach, CA, USA",  
year = {2019}  
}

@InProceedings{AIC22,
author = {M. Naphade and S. Wang and D. C. Anastasiu and Z. Tang and M. Chang and Y. Yao and L. Zheng and M. Shaiqur Rahman and A. Venkatachalapathy and A. Sharma and Q. Feng and V. Ablavsky and S. Sclaroff and P. Chakraborty and A. Li and S. Li and R. Chellappa},
title = "The 6th AI City Challenge",
booktitle = "2022 IEEE/CVF Conference on Computer Vision and Pattern Recognition Workshops (CVPRW)",
month = "June",
year = {2022},
pages = "3346-3355",
doi = "10.1109/CVPRW56347.2022.00378",
publisher = "IEEE Computer Society"
}

@inproceedings{InternImage,
  title={Internimage: Exploring large-scale vision foundation models with deformable convolutions},
  author={Wang, Wenhai and Dai, Jifeng and Chen, Zhe and Huang, Zhenhang and Li, Zhiqi and Zhu, Xizhou and Hu, Xiaowei and Lu, Tong and Lu, Lewei and Li, Hongsheng and others},
  booktitle={Proceedings of the IEEE/CVF Conference on Computer Vision and Pattern Recognition},
  pages={14408--14419},
  year={2023}
}

@article{intro2kotzenmacher2005evaluation,
  title={Evaluation of portable non-intrusive traffic detection system},
  author={Kotzenmacher, Jerry and Minge, Erik D and Hao, Bingwen},
  journal={IMSA Journal},
  volume={43},
  number={3},
  year={2005}
}

@inproceedings{yu2022dair,
	Author = {Yu, H. and Luo, Y. and Shu, M. and Huo, Y. and Yang, Z. and Shi, Y. and Guo, Z. and Li, H. and Hu, X. and Yuan, J. and others},
	Booktitle = {Proceedings of the IEEE/CVF Conference on Computer Vision and Pattern Recognition},
	Pages = {21361--21370},
	Title = {{DAIR-V2X}: A large-scale dataset for vehicle-infrastructure cooperative {3D} object detection},
	Year = {2022}}

@inproceedings{cress2022A9,
	Author = {Cre{\ss}, C. and Zimmer, W. and Strand, L. and Fortkord, M. and Dai, S. and Lakshminarasimhan, V. and Knoll, A.},
	Booktitle = {IEEE Intelligent Vehicles Symposium (IV)},
	Title = {A9-Dataset: Multi-Sensor Infrastructure-Based Dataset for Mobility Research},
	Year = {2022}}

@article{wen2020ua,
	Author = {Wen, L. and Du, D. and Cai, Z. and Lei, Z. and Chang, M. and Qi, H. and Lim, J. and Yang, M. and Lyu, S.},
	Journal = {Computer Vision and Image Understanding},
	Pages = {102907},
	Publisher = {Elsevier},
	Title = {{UA-DETRAC}: A new benchmark and protocol for multi-object detection and tracking},
	Volume = {193},
	Year = {2020}}

@article{ShokrolahShirazi2019,
	author = {S. Shirazi, M. and Morris, B. T.},
	da = {2019/09/01},
	doi = {10.1007/s00138-019-01040-w},
	id = {Shokrolah Shirazi2019},
	isbn = {1432-1769},
	journal = {Machine Vision and Applications},
	number = {6},
	pages = {1097--1109},
	title = {Trajectory prediction of vehicles turning at intersections using deep neural networks},
	ty = {JOUR},
	url = {https://doi.org/10.1007/s00138-019-01040-w},
	volume = {30},
	year = {2019}}

@article{ZHONG200725,
	author = {M. Zhong and G. Liu},
	journal = {Journal of Transportation Systems Engineering and Information Technology},
	number = {6},
	pages = {25-38},
	title = {Establishing and Managing Jurisdiction-wide Traffic Monitoring Systems: North American Experiences},
	volume = {7},
	year = {2007}}

@INPROCEEDINGS{9523074,
  author={Gloudemans, D. and Work, D. B.},
  booktitle={2021 IEEE/CVF Conference on Computer Vision and Pattern Recognition Workshops (CVPRW)}, 
  title={Fast Vehicle Turning-Movement Counting using Localization-based Tracking}, 
  year={2021},
  volume={},
  number={},
  pages={4150-4159},
  doi={10.1109/CVPRW53098.2021.00469}}

@INPROCEEDINGS{9522996,
  author={Ha, S. V. and Chung, N. M. and Nguyen, T. and Phan, H. N.},
  booktitle={2021 IEEE/CVF Conference on Computer Vision and Pattern Recognition Workshops (CVPRW)}, 
  title={Tiny-PIRATE: A Tiny model with Parallelized Intelligence for Real-time Analysis as a Traffic countEr}, 
  year={2021},
  volume={},
  number={},
  pages={4114-4123},
  doi={10.1109/CVPRW53098.2021.00465}}

@INPROCEEDINGS{9523000,
  author={Kocur, Viktor and Ftáčnik, Milan},
  booktitle={2021 IEEE/CVF Conference on Computer Vision and Pattern Recognition Workshops (CVPRW)}, 
  title={Multi-Class Multi-Movement Vehicle Counting Based on CenterTrack}, 
  year={2021},
  volume={},
  number={},
  pages={4004-4010},
  doi={10.1109/CVPRW53098.2021.00452}}

@INPROCEEDINGS{9523072,
  author={Tran, D. N. and Pham, L. H. and Nguyen, H. and Tran, T. Hu. and Jeon, H. and Jeon, J. W.},
  booktitle={2021 IEEE/CVF Conference on Computer Vision and Pattern Recognition Workshops (CVPRW)}, 
  title={A Region-and-Trajectory Movement Matching for Multiple Turn-counts at Road Intersection on Edge Device}, 
  year={2021},
  volume={},
  number={},
  pages={4082-4089},
  doi={10.1109/CVPRW53098.2021.00461}}

@INPROCEEDINGS{7045864,
  author={Wang, W. and Gee, T. and Price, J. and Qi, H.},
  booktitle={2015 IEEE Winter Conference on Applications of Computer Vision}, 
  title={Real Time Multi-vehicle Tracking and Counting at Intersections from a Fisheye Camera}, 
  year={2015},
  volume={},
  number={},
  pages={17-24},
  doi={10.1109/WACV.2015.10}}

@ARTICLE{7384599,
  author={S. Shirazi, M. and Morris, B. T.},
  journal={IEEE Intelligent Transportation Systems Magazine}, 
  title={Vision-Based Turning Movement Monitoring:Count, Speed \& Waiting Time Estimation}, 
  year={2016},
  volume={8},
  number={1},
  pages={23-34},
  doi={10.1109/MITS.2015.2477474}}

@INPROCEEDINGS{9562804,
  author={Shirazi, M. S. and Patooghy, A.},
  booktitle={2021 IEEE International Smart Cities Conference (ISC2)}, 
  title={A Vision-based Framework for Intersection Monitoring and Signal Evaluation}, 
  year={2021},
  volume={},
  number={},
  pages={1-4},
  doi={10.1109/ISC253183.2021.9562804}}

@INPROCEEDINGS{10444381,
  author={Nakano, K. and Nakazawa, M. and Zuzak, M.},
  booktitle={2024 IEEE International Conference on Consumer Electronics (ICCE)}, 
  title={Complementing Vehicle Trajectories Using Two Camera Viewpoints}, 
  year={2024},
  volume={},
  number={},
  pages={1-6},
  doi={10.1109/ICCE59016.2024.10444381}}

@INPROCEEDINGS{9150708,
  author={Abdelhalim, A. and Abbas, M.},
  booktitle={2020 IEEE/CVF Conference on Computer Vision and Pattern Recognition Workshops (CVPRW)}, 
  title={Towards Real-time Traffic Movement Count and Trajectory Reconstruction Using Virtual Traffic Lanes}, 
  year={2020},
  volume={},
  number={},
  pages={2527-2533},
  doi={10.1109/CVPRW50498.2020.00304}}

@INPROCEEDINGS{9151032,
  author={Yu, L. and Feng, Q. and Qian, Y. and Liu, W. and Hauptmann, A. G.},
  booktitle={2020 IEEE/CVF Conference on Computer Vision and Pattern Recognition Workshops (CVPRW)}, 
  title={Zero-VIRUS: Zero-shot Vehicle Route Understanding System for Intelligent Transportation}, 
  year={2020},
  volume={},
  number={},
  pages={2534-2543},
  doi={10.1109/CVPRW50498.2020.00305}}

@INPROCEEDINGS{9150961,
  author={Bui, K. N. and Yi, H. and Cho, J.},
  booktitle={2020 IEEE/CVF Conference on Computer Vision and Pattern Recognition Workshops (CVPRW)}, 
  title={A Vehicle Counts by Class Framework using Distinguished Regions Tracking at Multiple Intersections}, 
  year={2020},
  volume={},
  number={},
  pages={2466-2474},
  doi={10.1109/CVPRW50498.2020.00297}}

@INPROCEEDINGS{9150753,
  author={Wang, Z. and Bai, B. and Xie, Y. and Xing, T. and Zhong, B. and Zhou, Q. and Meng, Y. and Xu, B. and Song, Z. and Xu, P. and Hu, R. and Chai, H.},
  booktitle={2020 IEEE/CVF Conference on Computer Vision and Pattern Recognition Workshops (CVPRW)}, 
  title={Robust and Fast Vehicle Turn-counts at Intersections via an Integrated Solution from Detection, Tracking and Trajectory Modeling}, 
  year={2020},
  volume={},
  number={},
  pages={2598-2606},
  doi={10.1109/CVPRW50498.2020.00313}}

@INPROCEEDINGS{9151055,
  author={Chang, M. and Chiang, C. and Tsai, C. and Chang, Y. and Chiang, H. and Wang, Y. and Chang, S. and Li, Y. and Tsai, M. and Tseng, H.},
  booktitle={2020 IEEE/CVF Conference on Computer Vision and Pattern Recognition Workshops (CVPRW)}, 
  title={{AI} City Challenge 2020 – Computer Vision for Smart Transportation Applications}, 
  year={2020},
  volume={},
  number={},
  pages={2638-2647},
  doi={10.1109/CVPRW50498.2020.00318}}

@INPROCEEDINGS{9150979,
  author={Liu, Z. and Zhang, W. and Gao, X. and Meng, H. and Tan, X. and Zhu, X. and Xue, Z. and Ye, X. and Zhang, H. and Wen, S. and Ding, E.},
  booktitle={2020 IEEE/CVF Conference on Computer Vision and Pattern Recognition Workshops (CVPRW)}, 
  title={Robust Movement-Specific Vehicle Counting at Crowded Intersections}, 
  year={2020},
  volume={},
  number={},
  pages={2617-2625},
  doi={10.1109/CVPRW50498.2020.00315}}

@INPROCEEDINGS{9522714,
  author={Lu, J. and Xia, M. and Gao, X. and Yang, X. and Tao, T. and Meng, H. and Zhang, W. and Tan, X. and Shi, Y. and Li, G. and Ding, E.},
  booktitle={2021 IEEE/CVF Conference on Computer Vision and Pattern Recognition Workshops (CVPRW)}, 
  title={Robust and Online Vehicle Counting at Crowded Intersections}, 
  year={2021},
  volume={},
  number={},
  pages={3997-4003},
  doi={10.1109/CVPRW53098.2021.00451}}

@INPROCEEDINGS{9150586,
  author={OSPINA, A. and TORRES, F.},
  booktitle={2020 IEEE/CVF Conference on Computer Vision and Pattern Recognition Workshops (CVPRW)}, 
  title={Countor: count without bells and whistles}, 
  year={2020},
  volume={},
  number={},
  pages={2559-2565},
  doi={10.1109/CVPRW50498.2020.00308}}

@INPROCEEDINGS{10421880,
  author={Monteiro, M. L. and Bernardino, A. and Pinto, H. S.},
  booktitle={2023 IEEE 26th International Conference on Intelligent Transportation Systems (ITSC)}, 
  title={BYTECounter: Improving Vehicle Turning-Movement Counting}, 
  year={2023},
  volume={},
  number={},
  pages={3456-3462},
  doi={10.1109/ITSC57777.2023.10421880}}

@article{BARCELLOS20151845,
	author = {P. Barcellos and C. Bouvi{\'e} and F. L. Escouto and J. Scharcanski},
	doi = {https://doi.org/10.1016/j.eswa.2014.09.045},
	issn = {0957-4174},
	journal = {Expert Systems with Applications},
	number = {4},
	pages = {1845-1856},
	title = {A novel video based system for detecting and counting vehicles at user-defined virtual loops},
	url = {https://www.sciencedirect.com/science/article/pii/S095741741400606X},
	volume = {42},
	year = {2015}}

@INPROCEEDINGS{4290146,
  author={Bas, E. and Tekalp, A. M. and Salman, F. S.},
  booktitle={2007 IEEE Intelligent Vehicles Symposium}, 
  title={Automatic Vehicle Counting from Video for Traffic Flow Analysis}, 
  year={2007},
  volume={},
  number={},
  pages={392-397},
  doi={10.1109/IVS.2007.4290146}}

@article{XIA2016672,
	author = {Y. Xia and X. Shi and G. Song and Q. Geng and Y. Liu},
	doi = {https://doi.org/10.1016/j.sigpro.2014.10.035},
	issn = {0165-1684},
	journal = {Signal Processing},
	pages = {672-681},
	title = {Towards improving quality of video-based vehicle counting method for traffic flow estimation},
	url = {https://www.sciencedirect.com/science/article/pii/S016516841400499X},
	volume = {120},
	year = {2016}}

@INPROCEEDINGS{5457465,
  author={Dore, A. and Beoldo, A. and Regazzoni, C. S.},
  booktitle={2009 IEEE 12th International Conference on Computer Vision Workshops, ICCV Workshops}, 
  title={Multitarget tracking with a corner-based particle filter}, 
  year={2009},
  volume={},
  number={},
  pages={1251-1258},
  doi={10.1109/ICCVW.2009.5457465}}

@INPROCEEDINGS{4587551,
  author={Z. Kim},
  booktitle={2008 IEEE Conference on Computer Vision and Pattern Recognition}, 
  title={Real time object tracking based on dynamic feature grouping with background subtraction}, 
  year={2008},
  volume={},
  number={},
  pages={1-8},
  doi={10.1109/CVPR.2008.4587551}}

@INPROCEEDINGS{1640826,
  author={Porikli, F. and Tuzel, O. and Meer, P.},
  booktitle={2006 IEEE Computer Society Conference on Computer Vision and Pattern Recognition (CVPR'06)}, 
  title={Covariance Tracking using Model Update Based on Lie Algebra}, 
  year={2006},
  volume={1},
  number={},
  pages={728-735},
  doi={10.1109/CVPR.2006.94}}

@INPROCEEDINGS{4118800,
  author={Song, X. and Nevatia, R.},
  booktitle={2007 IEEE Workshop on Motion and Video Computing (WMVC'07)}, 
  title={Detection and Tracking of Moving Vehicles in Crowded Scenes}, 
  year={2007},
  volume={},
  number={},
  pages={4-4},
  doi={10.1109/WMVC.2007.13}}

@INPROCEEDINGS{1570925,
  author={Yokoyama, M. and Poggio, T.},
  booktitle={2005 IEEE International Workshop on Visual Surveillance and Performance Evaluation of Tracking and Surveillance}, 
  title={A Contour-Based Moving Object Detection and Tracking}, 
  year={2005},
  volume={},
  number={},
  pages={271-276},
  doi={10.1109/VSPETS.2005.1570925}}

@inproceedings{kmeansplusplus,
author = {Arthur, D. and Vassilvitskii, S.},
title = {k-means++: the advantages of careful seeding},
year = {2007},
isbn = {9780898716245},
publisher = {Society for Industrial and Applied Mathematics},
address = {USA},
booktitle = {Proceedings of the Eighteenth Annual ACM-SIAM Symposium on Discrete Algorithms},
pages = {1027–1035},
numpages = {9},
location = {New Orleans, Louisiana},
series = {SODA '07}
}
\end{document}